\documentclass{article}
\usepackage{ijcai22}

\usepackage{times}
\usepackage{soul}
\usepackage{url}
\usepackage[hidelinks]{hyperref}
\usepackage[utf8]{inputenc}
\usepackage[small]{caption}
\usepackage{graphicx}
\usepackage{amsmath}
\usepackage{amsthm}
\usepackage{booktabs}
\usepackage{algorithm}
\usepackage{algorithmic}
\newtheorem{definition}{Definition}

\usepackage{amssymb}
\newcommand{\refeq}[1]{Eq.~\eqref{#1}}
\newcommand{\refdef}[1]{Definition~\ref{#1}}

\newcommand{\set}[1]{\mathcal{#1}}
\newcommand{\pnorm}[1]{\lVert{#1}\rVert}

\newcommand{\classifierSet}{\set{H}}
\newcommand{\classifier}{\ensuremath{h}}
\newcommand{\setX}{\ensuremath{\set{X}}}
\newcommand{\setY}{\ensuremath{\set{Y}}}
\newcommand{\xorig}{\ensuremath{\vec{x}_{\text{orig}}}}

\newcommand{\x}{\ensuremath{\vec{x}}}

\newcommand{\ycf}{\ensuremath{y}_{\text{cf}}}
\newcommand{\xcf}{\ensuremath{\vec{x}_{\text{cf}}}}

\newcommand{\RN}{\mathbb{R}}
\newcommand{\dimsym}{d}
\DeclareMathOperator*{\dist}{{d}}

\newcommand{\explanation}{\ensuremath{\gamma}}
\newcommand{\setExplanations}{\ensuremath{\Gamma}}
\newcommand{\explanationRegularization}{\ensuremath{\theta}}
\newcommand{\explanationFidelity}{\ensuremath{\Psi}}
\newcommand{\regression}{\ensuremath{f}}
\newcommand{\threshold}{\ensuremath{\delta}}

\newcommand{\timeVar}{t}
\newcommand{\timeWinLen}{T}
\DeclareMathOperator*{\regularization}{\ensuremath{{\theta}}}
\DeclareMathOperator*{\loss}{{\ell}}

\title{One Explanation to Rule them All --- Ensemble Consistent Explanations}

\author{
Andr\'e Artelt$^{1,2}$
\and
Stelios Vrachimis$^{2}$
\and
Demetrios Eliades$^{2}$
\and
Marios Polycarpou$^{2}$
\and
Barbara Hammer$^1$
\affiliations
$^1$Bielefeld University \and
$^2$University of Cyprus
\emails
\{artelt.andre, vrachimis.stelios, eliades.demetrios, mpolycar\}@ucy.ac.cy \and bhammer@techfak.de
}

\begin{document}

\maketitle

\begin{abstract}
  Transparency is a major requirement of modern AI based decision making systems deployed in real world. A popular approach for achieving transparency is by means of explanations.
  A wide variety of different explanations have been proposed for single decision making systems. 
  In practice it is often the case to have a set (i.e. ensemble) of decisions that are used instead of a single decision only, in particular in complex systems. Unfortunately, explanation methods for single decision making systems are not easily applicable to ensembles -- i.e. they would yield an ensemble of individual explanations which are not necessarily consistent, hence less useful and more difficult to understand than a single consistent explanation of all observed phenomena.
  We propose a novel concept for consistently explaining an ensemble of decisions locally with a single explanation -- we introduce a formal concept, as well as a specific implementation using counterfactual explanations.
\end{abstract}

\section{Introduction}

Nowadays, transparency is a natural and broadly accepted requirement~\cite{gdpr} of AI based decision making systems. Usually, transparency is realized by providing explanations of the system's behavior~\cite{ExplainableArtificialIntelligence,ExplainingBlackboxModelsSurvey}.
While it is not yet completely understood what exactly makes up a ``good'' explanation~\cite{doshivelez2017rigorous,offert2017i}, a lot of different explanation methodologies~\cite{molnar2019} such as feature relevances/importances~\cite{FeatureImportance} and example based methods~\cite{CaseBasedReasoning} like contrasting explanations~\cite{CounterfactualWachter,CounterfactualReviewChallenges} have been developed.
Furthermore, explanation methods can be classified into global and local methods~\cite{molnar2019}: While global methods try to explain the entire system, local methods explain the system for a particular input or small region of input samples only.

To the best of our knowledge, existing methods focus on single decisions only -- i.e.\ they provide explanations of a single model and (if local) applied to a single input. In scenarios related to complex systems, we face a set or ensemble of decisions instead of a single function that must be explained. Examples are systems which expand in space -- i.e.\ spatial models connected to network data -- or in time -- i.e.\ decisions within time-dependent dynamic systems.
A highly relevant domain in the real world are water distribution networks where various quantities of interest in different regions of the network are monitored\footnote{E.g. looking for anomalies such as leakages and sensor faults that might cause water loss or water contamination.} by different systems.
These are typically based on the shared sensor data~\cite{leakage-perez,leakage-sanz}, hence the  monitoring system is given by an ensemble of decisions based on shared features -- we will address this application domain in Section~\ref{sec:experiments}.

In this work we propose a formalization how to provide a consistent explanation for an ensemble  of decisions or models -- i.e. computing one single explanation instead of  a separate and possibly disconnected explanation for each model.
More specifically, our contributions are as follows:
\begin{enumerate}
    \item We introduce the abstract concept of \textit{ensemble consistent explanations}, and we  provide a particular realization using counterfactual explanations.
    \item We empirically demonstrate the usefulness of ensemble consistent counterfactual explanations in sensor fault detection \& localization in water distribution networks.
\end{enumerate}

The remainder of this work is structured as follows:
First, we briefly review the necessary foundations in Section~\ref{[sec:relatedwork}. Next in Section~\ref{sec:ensembleconsistentexplanations}, we propose the abstract concept of \textit{ensemble consistent explanations} and  a specific realization thereof using counterfactual explanations, dubbed \textit{ensemble consistent counterfactual explanations}. Following up on this, we empirically evaluate the proposed technology in the real-world task of sensor fault identification in water distribution networks in Section~\ref{sec:experiments}, a domain where ensembles arise naturally due to the spatial nature of the task. Finally, we close our work with a summary and conclusion in Section~\ref{sec:conclusion}.

\section{Foundations}\label{[sec:relatedwork}

As mentioned in the introduction, there exist a wide variety of different explanation methodologies~\cite{molnar2019}. A prominent instance of contrasting explanations are
Counterfactual explanations (often just called \textit{counterfactuals}), which state a change to some features of a given input such that the resulting data point, called the counterfactual, causes a different behavior of the system than the original input does. Thus, one can think of a counterfactual explanation as a suggestion of actions that change the model's behavior. One reason why counterfactual explanations are so popular is that there exists evidence that explanations used by humans are often contrasting in nature~\cite{CounterfactualsHumanReasoning} -- i.e. people  ask questions like \textit{``What needs to be changed in order to observe a different outcome?''}.
%
In order to keep the explanation (suggested change) simple -- i.e.\ easy to understand -- an obvious strategy is to aim for a minimum change so that the resulting sample (counterfactual) is similar/close to the original sample. This is formalized via the following definition~\refdef{def:counterfactual}.
\begin{definition}[(Closest) Counterfactual Explanation~\cite{CounterfactualWachter}]\label{def:counterfactual}
Assume a prediction function $\classifier:
\setX\to \setY$ is given (where often $\setX=\RN^\dimsym$). A counterfactual $\xcf \in \setX$ for a given input $\xorig \in \setX$ is given as a solution of the following optimization problem:
\begin{equation}\label{eq:counterfactualoptproblem}
\underset{\xcf \,\in\, \setX}{\arg\min}\; \loss\big(\classifier(\xcf), \ycf\big) + C \cdot \regularization(\xcf, \xorig)
\end{equation}
where $\loss(\cdot)$ denotes a loss function, $\ycf$ the target prediction, $\regularization(\cdot)$ a penalty for dissimilarity of $\xcf$ and $\xorig$, and $C>0$ denotes a regularization strength.
\end{definition}
Such counterfactuals  are also called \textit{closest counterfactuals} 
\cite{karimi2021survey,CounterfactualComputationSurvey}.
Despite their popularity,  counterfactual explanations have some weaknesses. One is posed by a missing uniqueness: There can exist more than one valid counterfactual explanation -- this is called  Rashomon effect~\cite{molnar2019}. In such cases, it is not clear which or how many of those should be presented to the user. One common modeling approach is to ignore this effect, or to add a suitable regularization. If an ensemble of decisions should be explained, the Rashomon effect can yield inconsistent individual explanations, thus constitutes a severe problem.
Therefore, we propose a modeling which aims for consistency of such explanations.

\section{Ensemble Consistent Explanations}\label{sec:ensembleconsistentexplanations}
In the following, we assume that we are given a set (i.e.\ an ensemble) of decision functions $\classifier_i\in\classifierSet$, $\classifier_i:\setX\to\setY$ together with a sample $\x\in\setX$ -- as an example, this might be an ensemble in space or time.
We are looking for a single local explanation that consistently explains the behavior of all decision functions $\classifier_i(\cdot)$ for a given input state $\x$. 

We assume 
$\setExplanations(\x)$ refers to the set of all possible local explanation at a given input state $\x\in\setX$, which can be derived from any of the decisions within the ensemble. 
There exist two contrasting objectives, to choose a ``consistent optimum'' from this set: (i) how complex is an explanation (e.g.\ as measured by the distance to the original input state); we assume that $\explanationRegularization:\setExplanations\to\RN_{+}$ is a function that measures the complexity of a given explanation; smaller values correspond to a ``low complexity'' explanation while the opposite is true for ``high complexity'' explanation.
(ii) What is the fidelity of an explanation w.r.t.\ all decisions within the ensemble, i.e.\ does it
explain the behavior of every given function $\classifier_i(\cdot)$ at a given sample or not. We assume that  $\explanationFidelity:\setExplanations\times\classifierSet\times\setX\to\RN$ is a function that measures ``how well'' the explanation explains the behavior of a given function $\classifier_i(\cdot)$ at a given sample. We assume that $\explanationFidelity(\cdot)\geq 0$ holds if the explanation is ``correct'' (e.g.\ the computed value constitutes a counterfactual) and $\explanationFidelity(\cdot)< 0$ otherwise.

An \emph{ensemble-consistent explanation} is given by 
 a single, low complexity, local explanation $\explanation\in\setExplanations(\x)$ that explains the behavior/prediction of all given decision functions $\classifier_i(\cdot)$, i.e.
\begin{equation}\label{eq:ensembleconsistentexplanation:opt}
\begin{split}
&\underset{\explanation\,\in\,\setExplanations(\x)}{\min}\,\explanationRegularization(\explanation) \\
\text{s.t. } & \explanationFidelity(\explanation,\classifier_i,\x) \geq 0 \quad \forall\,i
\end{split}
\end{equation}
In practice, this problem needs to be relaxed to guarantee at least one  feasible solutions. Therefore we use slack variables $\xi_i$ that allow constraint violations,  which are weighted by a penalty $\lambda>0$:
\begin{equation}\label{eq:ensembleconsistentexplanation:opt:relaxed}
\begin{split}
&\underset{\explanation\,\in\,\setExplanations(\x)}{\min}\,\explanationRegularization(\explanation) + \lambda\sum_i \xi_i \\
\text{s.t. } & \explanationFidelity(\explanation,\classifier_i,\x) \geq 0 - \xi_i \quad \forall\,i\\
& \xi_i \geq 0 \quad \forall\,i
\end{split}
\end{equation}

\subsection{Consistent Counterfactual Explanations}
In the following, we use counterfactual explanations (see~\refdef{def:counterfactual} in Section~\ref{[sec:relatedwork}) as a specific type of local explanations for realizing the idea of ensemble consistent explanations as described in the previous section.

Because we consider a set of prediction functions $\classifier(\cdot)_i$, we might request potentially different target predictions ${\ycf}_i$ in the counterfactual explanation.

\subsubsection{Classification}
Assuming a set of binary classifiers $\classifier_i:\RN^\dimsym\to\{-1,1\}$. We can implement the idea of ensemble consistent explanations by refining~\refeq{eq:ensembleconsistentexplanation:opt:relaxed} as follows:
\begin{equation}
\begin{split}
&\underset{\xcf\,\in\,\RN^\dimsym}{\min}\,\regularization(\xcf, \x) + \lambda\sum_i \xi_i \\
\text{s.t. } & \classifier_i(\xcf)\cdot{\ycf}_i \geq 0 - \xi_i \quad \forall\,i\\
& \xi_i \geq 0 \quad \forall\,i
\end{split}
\end{equation}

\subsubsection{Regression}
Similar, for a set of regression functions $\regression_i:\RN^\dimsym\to\RN$, we refine~\refeq{eq:ensembleconsistentexplanation:opt:relaxed} as follows:
\begin{equation}\label{eq:eccf:regression}
\begin{split}
&\underset{\xcf\,\in\,\RN^\dimsym}{\min}\,\regularization(\xcf, \x) + \lambda\sum_i \xi_i \\
\text{s.t. } & \dist\big(\regression_i(\xcf) - {\ycf}_i\big) \leq \threshold_i + \xi_i \quad \forall\,i\\
& \xi_i \geq 0 \quad \forall\,i
\end{split}
\end{equation}
where $\dist:\RN\to\RN_{+}$ denotes a function for penalizing prediction errors -- e.g. the absolute error or the squared error\footnote{$d(a)=a^2$ or $d(a)=|a|$} -- and $\threshold_i \geq 0$ denotes a tolerance threshold on the prediction error of the $i$-th regression function, since it might be numerically difficult to achieve exactly ${\ycf}_i$ as a prediction.

\section{Experiments}\label{sec:experiments}

We empirically evaluate our proposed \textit{ensemble consistent counterfactual explanations} in the important real world application of sensor fault detection and localization in water distribution networks (WDNs) -- in this context, we use our proposed explanations for identifying a faulty sensor.
The implementation of the experiments is publicly available\footnote{\url{https://github.com/andreArtelt/EnsembleConsistentExplanations}}.

\subsection{Setup}
\begin{figure}
    \centering
    \includegraphics[width=0.5\textwidth]{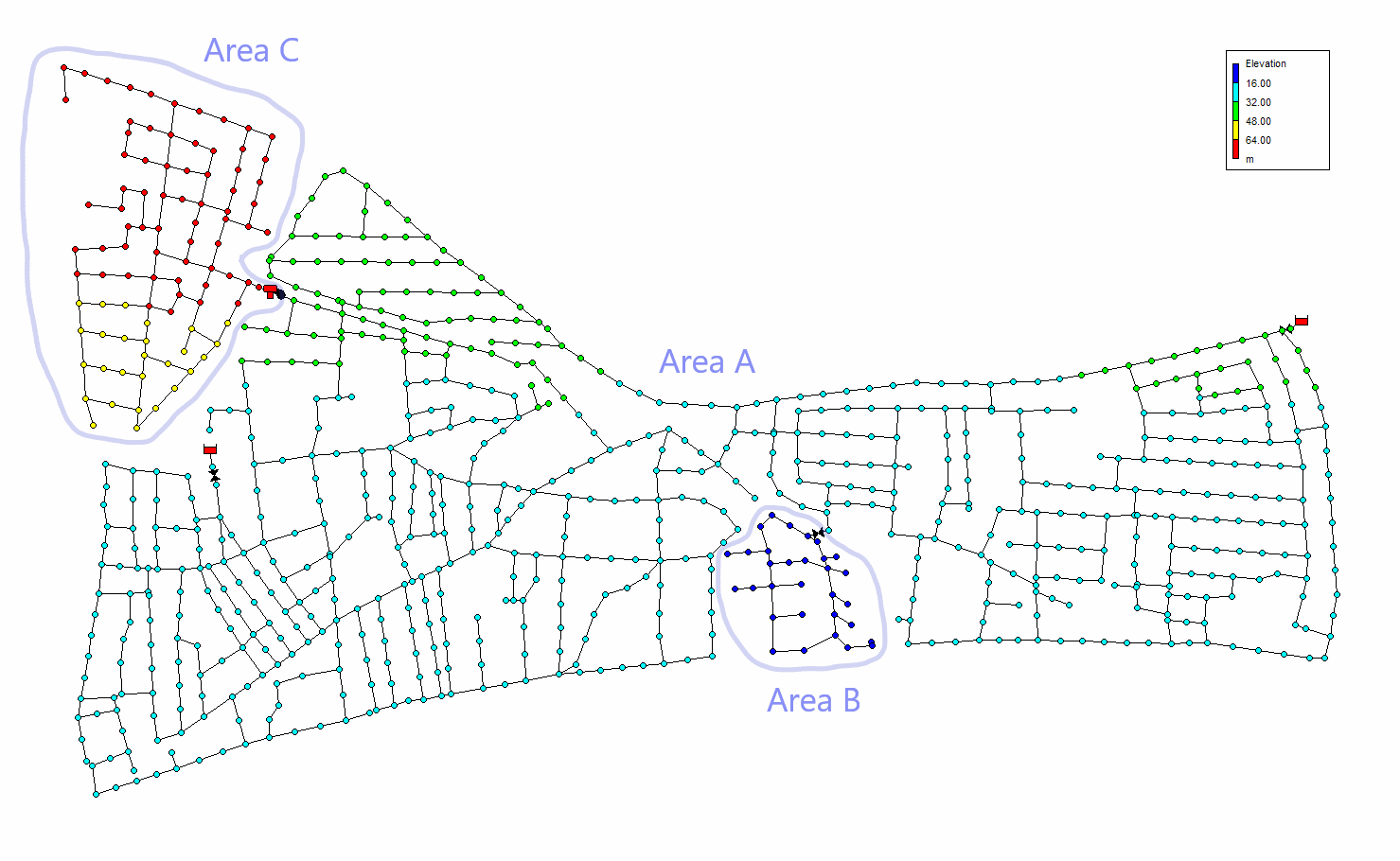}
    \caption{L-Town network -- we only use the hydraulically isolated ``Area A'', where $29$ pressure and $2$ flow sensors are installed.}
    \label{fig:l_town}
\end{figure}
\begin{figure}
    \centering
    \includegraphics[width=0.5\textwidth]{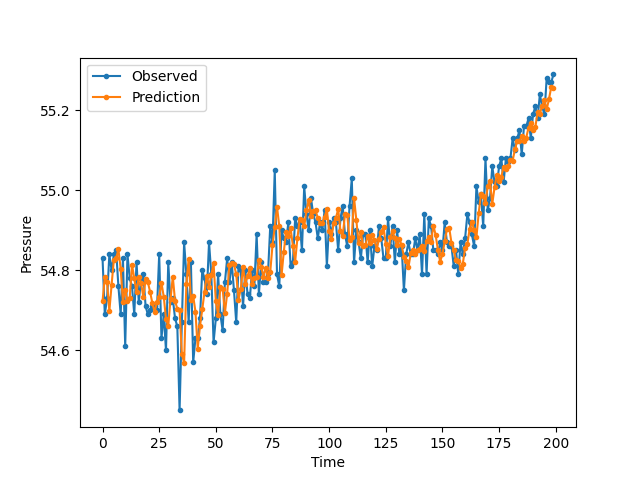}
    \caption{Illustration of the a virtual pressure sensor: Predicted vs. observed pressure values at node \textit{n549}.}
    \label{fig:l_town:model}
\end{figure}
As a WDN, we use a variation of the popular L-Town water distribution network~\cite{vrachimis2020battledim} -- 
for simplicity, we demonstrate our approach on the hydraulically isolated Area A, by closing the valves that provide water to Area B and C (see Fig.~\ref{fig:l_town}). Area A consists of $661$ nodes and $766$ links. The benchmark assumes the existence of $29$ pressure sensors at nodes (optimally placed to maximize the detectability of leakages) and $2$ flow sensors at the water inlets of the network.
We implement a residual based anomaly detection method by means of virtual sensors~\cite{eliades2012leakage,santos2019estimation,vaquet2022icann}. We build a virtual sensor $f_i:\RN^{30} \to \RN$ for each of the $29$ pressure sensors -- the virtual sensor tries to predict the pressure at its specific location based on the previous measurements from all other $28$ (i.e.\ $29 - 1$) pressure and $2$ flow sensors (see Fig.~\ref{fig:l_town:model}). The virtual sensors are implemented using linear regression, we use a sliding window of size $\timeWinLen=3$ time steps for averaging the other past sensor values which are then fed into the linear regression:
\begin{equation}
    f_i(\vec{x}^{i}_{\timeVar-1}) = \vec{w}_i^\top\vec{x}^{i}_{\timeVar-1} + b_i \;\text{with}\; \vec{x}^{i}_{\timeVar-1} = \frac{1}{\timeWinLen}\sum_{j=1}^{\timeWinLen} (\vec{x}_{\timeVar-j})_{\neq i}
\end{equation}
where $(\vec{x}_{\timeVar-j})_{\neq i}$ denotes the sensor measurements at time $\timeVar - j$ at all nodes except the $i$-th node.
The anomaly detection itself is realized using a simple threshold $\delta$ that is estimated during training on a fault free time window -- we raise an alarm whenever any predicted pressure $f_i(\vec{x}^{i}_{\timeVar-1})$ (from the virtual sensor) deviates from the observed pressure ${(\vec{y}_{\timeVar})}_{i}$:
\begin{equation}\label{eq:anomalydetection}
\pnorm{f_i(\vec{x}^{i}_{\timeVar-1}) - (\vec{y_\timeVar})_{i}}_{\infty} > \delta
\end{equation}

We generate $434$ scenarios using the variation of the L-Town network where we place sensor faults of different magnitude on the pressure sensors -- we consider three different magnitudes for each of the following five types of sensor faults~\cite{sensorfaultbook}:
\begin{enumerate}
    \item A constant offset of the sensors measurement compared to the true measured quantity.
    \item Gaussian noise added to the sensor measurements.
    \item Sensor power failures, which results to the measurement being equal to zero.
    \item An offset of the sensor measurement, linearly proportional to the true measured quantity.
    \item Sensor drift, resulting in an increasing measurement value until a maximum limit is reached.
\end{enumerate}
Each scenario is exactly three month long and contains exactly one sensor fault -- the first month is always fault free, so it can be used for training the virtual sensors.

\subsection{Results}
\begin{table}
\centering
\begin{tabular}{lrr}
\toprule
Metric  & Score (avg. $\pm$ variance) \\
\midrule
True Positives & $ 1.0000 \pm 0.0000$ \\
True Negatives & $ 0.9951 \pm 0.0009$ \\
False Positives & $ 0.0000 \pm 0.0000$ \\
False Negatives & $ 0.0049 \pm 0.0009$ \\
Detection delay & $ 1.0021 \pm 0.0020$ 
\end{tabular}
\caption{Evaluation of anomaly detection}
\label{tab:anomalydetection:eval}
\end{table}
First, we evaluate the performance of the implemented residual based anomaly detection by reporting the mean and variance of the confusion matrix and detection delay (time until the first alarm is raised) over all scenarios -- the results are shown in Table~\ref{tab:anomalydetection:eval}. We observe that our implemented anomaly detector is consistently able to detect all sensor faults with an average delay of $1$-$2$ time steps.
\begin{figure}
    \centering
    \includegraphics[width=0.5\textwidth]{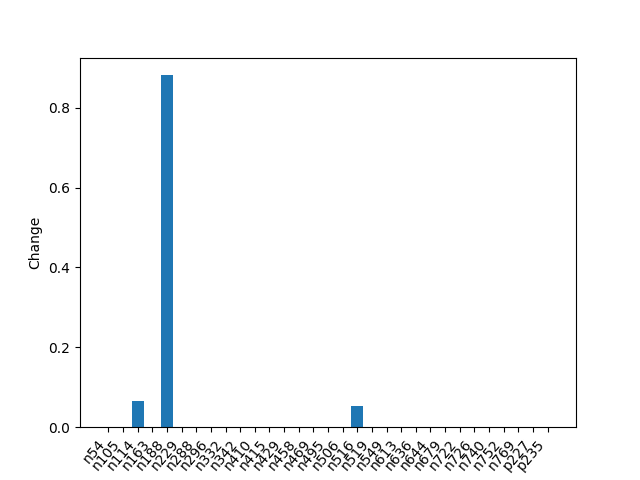}
    \caption{Distribution of predicted faulty sensors according to the independent counterfactuals for a raised alarm in case of a sensor fault at node \textit{n332} -- note that the predictions are wrong.}
    \label{fig:counterfactual_fingerprint:baseline}
\end{figure}
\begin{figure}
    \centering
    \includegraphics[width=0.5\textwidth]{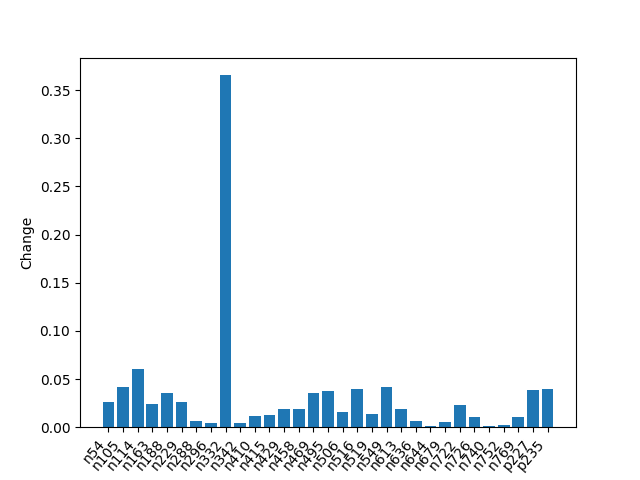}
    \caption{Illustration of a normalized \textit{ensemble consistent counterfactual explanations} for a raised alarm in case of a sensor fault at node \textit{n332}.}
    \label{fig:counterfactual_fingerprint}
\end{figure}
\begin{table}
\centering
\begin{tabular}{lrr}
\toprule
Metric  & Score (avg. $\pm$ variance) \\
\midrule
Our method   & $0.97 \pm 0.03$   \\
Baseline     & $0.00 \pm 0.00$
\end{tabular}
\caption{Evaluation (in \%) of successful sensor fault localization.}
\label{tab:faultlocalization:eval}
\end{table}
Next, we evaluate the fidelity of our proposed \textit{ensemble consistent counterfactual explanations} by trying to identify the faulty sensor from the explanation -- i.e.\ identifying the cause of the observed anomaly.
For every detected fault (i.e.\ raised alarm), we compute an ensemble consistent counterfactual explanation by solving~\refeq{eq:eccf:regression} using convex programming and normalize it for convenience -- see Fig.~\ref{fig:counterfactual_fingerprint} for an example. That is we are looking for a single explanation that tells us how to change the sensor measurements such that no alarm would have been raised.
Note that since an anomaly stays active for some time, we have a sequence of permanent alarms.
We select the sensor with the largest magnitude of proposed changes (according to the explanation) as our prediction of the faulty sensor which we then compare with the known ground truth.
Data and predictions are a bit noisy (see Fig.~\ref{fig:l_town:model}), hence other sensor components  might be non-vanishing as well, as can be seen in Fig.~\ref{fig:counterfactual_fingerprint}.
We compare the consistent explanation to the baseline which is obtained by aggregating independent 
local counterfactual explanations for every decision:
that means, we compute a counterfactual explanations of each model separately, and select the sensor with the largest change as an estimate of the faulty sensor -- we aggregate these estimates by selecting the most common estimate of the faulty sensor.
The results are shown in Table~\ref{tab:faultlocalization:eval}. We observe that, in contrast to the baseline which is not capable of identifying the faulty sensor, 
we are able to correctly identify the faulty sensor in almost every scenario. This fact demonstrates the high fidelity and usefulness of our proposed explanation methodology in this application example.
It turns out that the independent counterfactuals are very sparse but almost always have the peak at the same (wrong) sensor (see Fig.~\ref{fig:counterfactual_fingerprint:baseline} for an example) -- i.e. they exploit some model specific properties to get a sparse counterfactual, which is unrelated to the faulty sensor.

\section{Summary \& Conclusion}\label{sec:conclusion}

In this work we proposed the concept of \textit{ensemble consistent explanations} for computing explanations of an ensemble of decisions instead of a single decision only. Ensembles naturally occur for processes  with spacial distribution -- i.e.\ a set of models such as virtual sensors in networks as we considered in our experiments -- or in time -- i.e.\ different instances of a single model evolving over time.
Furthermore, we also proposed a specific implementation of this concept using counterfactual explanations yielding \textit{ensemble consistent counterfactual explanations}.
For the purpose of an empirical evaluation, we applied our proposed explanations to the (important real-world) problem of sensor fault detection and localization in water distribution networks. We observed that our proposed explanations show very good performance for explaining the detected anomalies and identifying the faulty sensors which caused the observed anomaly.

A couple of extensions and  directions of further research can be based on this initial work.
In particular, driven by the specific application domain, we aim for the following:
How does the method perform for more than one sensor fault, i.e.\ explanations which need to identify multiple features or more complex signals?
How is the behavior for more complex nonlinear dynamics modeled by deep networks?
What are suitable extensions to temporal dynamics and ensembles in time rather than space, such as tackled using deep recurrent models or attention mechanisms?
Are there formal guarantees to substantiate consistency of the proposed method? 


\section*{Acknowledgments}

We gratefully acknowledge funding from the VW-Foundation for the project \textit{IMPACT}, and funding from from the European Research Council (ERC) under the ERC Synergy Grant Water-Futures (Grant agreement No. 951424).

\bibliographystyle{named}
\bibliography{ijcai22}

\end{document}